\documentclass{article}
\usepackage{arxiv}
\usepackage{nicefrac}       
\usepackage{microtype}      
\usepackage{booktabs}       
\usepackage{natbib}
\usepackage{doi}
\usepackage{breqn}
\usepackage{overpic}
\usepackage{adjustbox}
\usepackage{amsmath}
\usepackage{amssymb}
\usepackage{float}
\usepackage[T1]{fontenc}
\usepackage{graphicx}
\usepackage{caption}
\usepackage{subcaption}
\usepackage[utf8]{inputenc}
\usepackage{mathtools}
\usepackage{multicol}
\usepackage{multirow}
\usepackage{wasysym}

\title{Mathematical Framing for Different Agent Strategies}
\author{
    \textbf{Philip Stephens} \qquad \textbf{Emmanuel Salawu} \\
    \\
    Google Cloud AI
}

\renewcommand{\shorttitle}{Mathematical Framing for Different Agent Strategies}

\hypersetup{
pdftitle={Mathematical Framing for Different Agent Strategies},
pdfsubject={q-bio.NC, q-bio.QM}, 
pdfauthor={Philip Stephens, Emmanuel Salawu},
pdfkeywords={AI Agents, Agent Strategies, Multi-Agent Systems, Mathematical Framing for Agents},
}

\begin{document}
\maketitle

\begin{abstract}
We introduce a unified mathematical and probabilistic framework for understanding and comparing diverse AI agent strategies. We bridge the gap between high-level agent design concepts, such as ReAct, multi-agent systems, and control flows, and a rigorous mathematical formulation. Our approach frames agentic processes as a chain of probabilities, enabling a detailed analysis of how different strategies manipulate these probabilities to achieve desired outcomes. Our framework provides a common language for discussing the trade-offs inherent in various agent architectures. One of our many key contributions is the introduction of the "Degrees of Freedom" concept, which intuitively differentiates the optimizable levers available for each approach, thereby guiding the selection of appropriate strategies for specific tasks. This work aims to enhance the clarity and precision in designing and evaluating AI agents, offering insights into maximizing the probability of successful actions within complex agentic systems.
\end{abstract}

\keywords{ AI Agents \and Agent Strategies \and Multi-Agent Systems \and Mathematical Framing for Agents}

\section{Introduction}

The proliferation of sophisticated Large Language Models (LLMs) has catalyzed a paradigm shift in AI, moving beyond static text generation towards the development of autonomous, goal-directed AI agents  \cite{Zhiheng_Xi_et_al_2025} \cite{Prerak_Garg_Divya_Beeram2024}. These agents (which are capable of reasoning, planning, and interacting with digital environments through tools) represent a significant step towards more general artificial intelligence  \cite{Alhassan_Mumuni_Fuseini_Mumuni2025}. An agent, in this context, can be understood as a system that perceives its environment, formulates a plan through a series of reasoning steps (or "thoughts"), and executes actions to achieve a specific objective \cite{Stuart_Jonathan_Russell__and__Peter_Norvig_1995}. This fundamental capability has unlocked a vast array of applications, from automated software engineering and complex data analysis to personalized digital assistants and scientific discovery  \cite{Guannan_Liang_Qianqian_Tong2025}.

Further complicating this landscape are the myriad techniques for managing the agent's process. These include "control flow" mechanisms, which partition the action space and guide the agent through a predefined graph or state machine, thereby constraining its behavior to improve reliability  \cite{Shunyu_Yao_et_al_2023}. This stands in contrast to the "random walk" nature of a basic ReAct loop, which, while flexible, can suffer from a lack of convergence or fall into hallucinatory loops  \cite{Mert_Cemri_et_al_2025}. Moreover, the very "context" provided to the agent – the initial prompt, the available tools, and the method of updating its state with new observations – is a critical design element. This "context engineering" can be static (prompt engineering) or dynamic, where the state is strategically manipulated, summarized, or selectively updated at each step  \cite{Sander_Schulhoff_et_al_2024} \cite{Pranab_Sahoo_et_al_2024} \cite{Joon_Sung_Park_et_al_2023}.

This divergence in design philosophy – from simple loops to complex graphs, from single agents to collaborative swarms – has created a significant conceptual and practical challenge. While empirical benchmarks can compare the end-to-end performance of different agents on specific tasks  \cite{Mohit_Shridhar_et_al_2020} \cite{Xiao_Liu_et_al_2023}, there exists a profound gap in our formal understanding of these systems. \textbf{There is a lack of a unified mathematical language to describe, analyze, and compare these diverse strategies.} \textit{How, for instance, can we rigorously quantify the trade-offs between a complex, multi-step "deep thinking" inference model and a multi-agent system that partitions the problem? How do we mathematically frame the intuitive benefit of "collaboration" or "negotiation" between agents?} \textbf{Without such a framework, the design of AI agents remains a high-dimensional, empirical art, lacking the precise engineering principles needed for systematic optimization and an understanding of the trade-offs inherent in any given architectural choice.}

This paper bridges this gap. We posit that the behavior of any agentic system, regardless of its architecture, can be fundamentally understood as a probabilistic process. The agent's ultimate goal is to maximize the probability of taking a specific sequence of actions (the "goal actions") that leads to a successful outcome, given some initial context. \textbf{By framing the agentic processes as a chain of probabilities – a formulation that naturally lends itself to the mathematics of Markov chains – we can begin to formally dissect and compare these seemingly disparate strategies.} This paper introduces such a unified mathematical and probabilistic framework, providing a common language to analyze the levers available to the agent designer. We introduce the concept of "Degrees of Freedom" – the distinct optimizable parameters within this probabilistic system – to intuitively differentiate the levers available in each approach, from simple prompt engineering to complex inter-agent collaboration.

\section{Prior Work}

The concepts presented in this paper build upon several distinct but converging lines of research in artificial intelligence, machine learning, and computational theory.

\subsection{Chain-of-Thought and Reasoning-as-Action}

A foundational precursor to modern agentic systems is the "Chain-of-Thought" (CoT) prompting technique  \cite{Jason_Wei_et_al_2022} \cite{Bingxi_Zhao_et_al_2025}. CoT demonstrated that eliciting intermediate reasoning steps from an LLM before arriving at a final answer dramatically improves performance on complex logical, mathematical, and symbolic reasoning tasks  \cite{Jason_Wei_et_al_2022}. This insight – that the generation of "thoughts" is a critical component of the problem-solving process – was formalized and externalized in the ReAct framework  \cite{Shunyu_Yao_et_al_2023}. ReAct proposed a simple yet powerful loop where the model generates both a "thought" (a reasoning trace) and an "action" (an executable command, such as a search query or code execution). The observation from this action is then fed back into the model's context, informing the next thought-action pair. This paper heavily references the ReAct formalism, using it as a canonical example of a monolithic agent. Our probabilistic framing models the ReAct loop as a specific type of Markov chain, where the key insight is that the generation of thoughts, \( t_i \) , serves to increase the probability of selecting the correct action, \( a_i \) , from a given state, \( s_{i-1} \) . Prior work has largely focused on the empirical success of ReAct and its variants, rather than its formal probabilistic properties and limitations, which this paper explores in depth. 

\subsection{Multi-Agent Systems (MAS)}

The concept of Multi-Agent Systems (MAS) has a rich history in distributed artificial intelligence, long predating the LLM era  \cite{Peter_Stone_Manuela_Veloso2000}. Classical MAS research focused on the coordination, collaboration, and negotiation strategies of multiple autonomous, often symbolic, agents to solve problems beyond the capability of any single agent. This body of work provides foundational concepts for inter-agent communication protocols, task decomposition, and coalition formation  \cite{Tuomas_Sandholm_et_al_1999}. In the modern context, LLM-based multi-agent systems have re-emerged as a powerful paradigm  \cite{Joon_Sung_Park_et_al_2023}. These systems typically involve multiple LLM-powered agents assuming different "roles" (e.g., "coder," "tester," "project manager") and communicating via natural language to complete a complex task  \cite{Chen_Qian_et_al_2024}. While empirically powerful, this work often lacks a rigorous mathematical description of why such systems are effective. This paper contributes a novel formulation by modeling the inter-agent communication itself as a probabilistic element. We introduce probabilities like \( P(c_L | \mathbf{a}_L) \)  – the probability of generating a specific context \( c_L \)  for another agent, given the actions \( \mathbf{a}_L \)  of the first – to formally represent the new "Degrees of Freedom" that collaboration and negotiation introduce into the optimization landscape. 

\subsection{Control Flow and Structured Reasoning}

A significant body of work has explored methods to move beyond the "random walk" limitation of simple ReAct loops by imposing more structure on the agent's reasoning process. This includes "Tree-of-Thoughts," which explores multiple reasoning paths in parallel and uses pruning or search algorithms to select the most promising one  \cite{Shunyu_Yao_et_al_2023} \cite{Aske_Plaat_et_al_2025}. Other approaches have explicitly utilized graph-based structures, where the agent's state transitions are constrained to the nodes and edges of a predefined graph, representing a formal control flow. These methods can be seen as dynamically manipulating the set of available actions or the state update function at each step to enforce constraints and improve the probability of reaching a goal state. Our framework subsumes these approaches by describing them as a strategy of modifying the probability kernel \( P(a_i | s_{i-1}) \)  at each step \( i \) . This manipulation, whether by changing the prompt (and thus the state \( s \) ), the available tools (constraining the action space \( \mathbf{a} \) ), or the inference functional \( \mathcal{F} \)  itself, is a method of partitioning the action space to increase the probability of the desired sub-actions. 

\subsection{Formal Models of Agent Behavior}

There have been prior efforts to apply formal methods, such as those from control theory and reinforcement learning  \cite{R.S._Sutton_A.G._Barto1998}, to agentic systems. Markov Decision Processes (MDPs) and their variants are a common tool for modeling goal-directed behavior in an environment  \cite{Martin_L._Puterman1994} \cite{Matthijs_T._J._Spaan2012}. However, in the context of LLM-based agents, the state space is often intractably large (the space of all possible text contexts) and the transition probabilities (the LLM's outputs) are implicit and non-stationary. While reinforcement learning techniques like finetuning have been used to optimize agent policies (effectively modifying the core model \( m \) of which \( \mathcal{F} \) is a functional over)  \cite{David_Ha2019}, these approaches do not typically provide a framework for comparing the architectural trade-offs of different agent designs (e.g., ReAct vs. Multi-Agent). \textbf{This paper provides a higher-level probabilistic formulation that is not concerned with training the underlying model parameters, but rather with analyzing the structural and probabilistic consequences of different \textit{architectural choices}.} By framing the objective as maximizing \( P(\mathbf{a}_g | c) \) , we provide a clear mathematical target that allows for the comparison of how different strategies (ReAct, Control Flow, Multi-Agent, model training, prompt engineering, etc.) create and exploit different "Degrees of Freedom" to achieve this maximization. This work is a bridge, connecting the high-level, conceptual design of agent systems with a rigorous, unified mathematical language.

\subsection{Key Contributions}

We provide a unified mathematical and probabilistic framework for the analysis and comparison of diverse AI agent strategies. Our primary contribution is the formulation of agentic processes as a chain of probabilities, \( P(a|s) \) , which allows for a precise examination of how different architectures and techniques manipulate these probabilities to maximize the likelihood of a successful action sequence, \( \mathbf{a}_g \) .

The key conceptual tools introduced are:

\subsubsection{Unified Probabilistic Formulation}

We formally define an agent \( A_{\mathcal{F},u}(\mathbf{a}|c) \)  as a probabilistic function, characterized by its inference functional \( \mathcal{F} \) , its state update function \( u \) , and its initial state \( s_0(c) \) . This allows us to place disparate strategies like ReAct, Control Flow, and Multi-Agent systems within a common mathematical context. 

\subsubsection{The "Degrees of Freedom" Concept}

We use this formalism to identify and differentiate the key "levers" or "knobs" that a designer can optimize for each strategy. For example, a ReAct loop offers optimization via prompt engineering (\( s_0(c) \) ) and the state update (\( u \) ), while a multi-agent system introduces entirely new probabilistic terms related to inter-agent collaboration, \( P(c_K | \mathbf{a}_K) \) , representing new surfaces for optimization. 

\subsubsection{Formalization of Collaboration}

We provide, to our knowledge, one of the first formal mathematical treatments of inter-agent collaboration and negotiation within an LLM agent framework. We model this as a probabilistic search for an optimal context, \( c_L \) , to pass between agents, thereby dynamically tuning the system's outcome probabilities without requiring model retraining. 

To address the practical realities of such systems, we also extend this framework to include the concept of "Collaboration Costs." We introduce a regularized objective function, \( \text{Maximize}( P(...) - \lambda \cdot \text{CollabCost}(\cdot) ) \) , which balances the drive to maximize outcome probability with the practical costs of latency, computation, and complexity inherent in inter-agent communication. This provides a more robust and realistic objective for designing agent systems that are not just effective, but also efficient.

\section{Probabilistic Framing}

We begin by providing a probabilistic framing for agentic operations. In general this can be defined as a chain of probabilities where each one represents the probability of a specific \textbf{action, \( a_i \) } occurring from a given \textbf{state, \( s_{i-1} \) } based on some input \textbf{context \( c \) }. I.e.

\begin{equation}
\label{eq:nolabel}
P(\mathbf{a} | c) = P(a_n|s_{n-1}, c)P(a_{n-1}|s_{n-2}, c)\dots P(a_1|s_0, c)
\end{equation}

If state \( s_i \)  is only a function of \( s_{i-1} \)  and \( a_i \) , this is a Markov chain. We can further say that there exists some \textbf{state update function \( u \) } such that \( s_i = u(a_i, a_{i-1},...a_1, s_{i-1}, s_{i-2},..., s_0) \)  (where in a Markov Chain it would be  \( s_i = u(a_i, s_{i-1}) \) ). Finally we can define an action \( a_i \)  as being composed of some choice of \textbf{action class, \( \alpha_i \) } with some argument \( x_i \)  such that the result of the action is the \textbf{observation \( o_i \) }  e.g. $\alpha_i(x_i) = o_i $. This is short handed with $a_i $ throughout where $a_i \to \{ \alpha_i, x_i, o_i \} $.

\begin{figure} 
\centering
\begin{subfigure}{.5\textwidth}
\centering
\includegraphics[width=4.34cm,height=6.79cm]{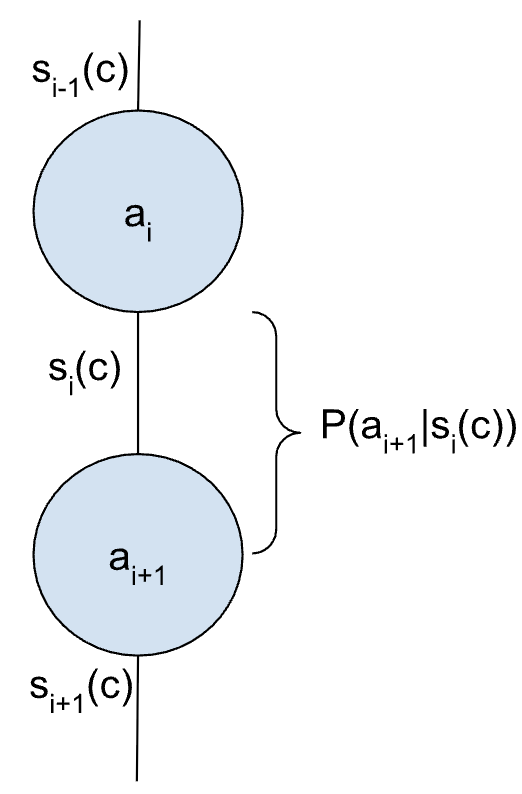}
\caption{}
\label{fig:graphical_representation_single_link_agentic}
\end{subfigure}%
\begin{subfigure}{.5\textwidth}
\centering
\includegraphics[width=9.03cm,height=8.49cm]{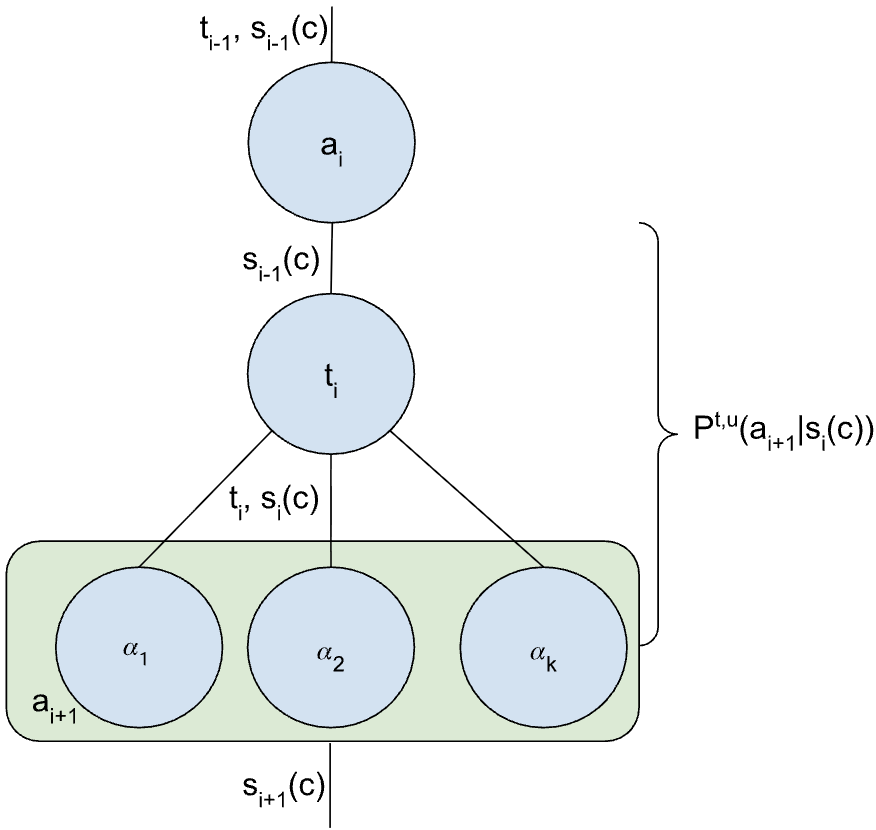}
\caption{}
\label{fig:probabilistic_expansion_action_step_incorporating}
\end{subfigure}
\caption{Graphical representation of a link in the probability chain. (a) Shows a single link in the chain, while (b) shows the action \( a_i+1 \) expanded into the class of possible actions \( \{\alpha\} \) as well as showing the inclusion of an explicit reasoning step.}
\end{figure}

A link in this chain can be graphically represented as shown in Fig. \ref{fig:graphical_representation_single_link_agentic}. A class of actions \( \{ \alpha \}\) expanded for a single \( a_{i+1} \) and also including a reasoning step, \( t_i \) is shown in Fig. \ref{fig:probabilistic_expansion_action_step_incorporating}.

\subsection{ReAct}

For the most commonly used agentic approach, the ReAct style prompting (wherein we consider the ReAct-style prompts and the tools afforded the agent are part of content, \( c \) , upon which the state, \( s_0 \) , depends), can be expressed as:

\begin{equation}
\label{eq:nolabel_1}
P^{t}(a_i|s_{i-1}) = P(a_i|t_i, s_{i-1})P(t_i|s_{i-1})
\end{equation}

for intermediate states, \( t_i \)  (thoughts) and is annotated by the superscript \( t \) . Thus the full probability of taking actions \( \mathbf{a} = \{a_1, a_2, ..., a_n\} \)  (and \( a_n \)  represents a terminal action) with internal thoughts \( \{t_1, t_2, ..., t_n\} \)  is

\begin{equation}
\label{eq:nolabel_2}
P^{t,u}(\mathbf{a}; \mathbf{t}|c) = P(a_n|t_n, s_{n-1})P(t_n|s_{n-1}) P(a_{n-1}|t_{n-1}, s_{n-2})P(t_{n-1}|s_{n-2}) ... P(a_1|t_1, s_0)P(t_1|s_0(c)),
\end{equation}

with update function

\begin{equation}
\label{eq:nolabel_3}
s_i = u(a_i, a_{i-1}, ..., a_1, t_n, ..., t_1, s_{i-1}, ..., s_0, c, \alpha, \beta, ...).
\end{equation}

for an arbitrary parameterization \( \alpha, \beta, ... \) .

The probability of taking actions \( \mathbf{a} \)  regardless of the internal thoughts is then (treating the thoughts as a continuous space)

\begin{equation}
\label{eq:nolabel_14}
P^{\text{ReAct}(t,u)}(\mathbf{a}|c) = \int dt_1 ... \int dt_n P^{t,u}(\mathbf{a}, \mathbf{t} | c).
\end{equation}

It is important to note that (where \( \odot \)  represents an identity functional, e.g. just a raw LLM with function calling)

\begin{equation}
\label{eq:nolabel_4}
P^{\text{ReAct}(t,u)}(\mathbf{a}|c) \neq P^{\odot,u}(\mathbf{a}|c)
\end{equation}

which is the key insight behind the ReAct formalism - generating the thoughts increases the likelihood of the right actions. 

In general the ultimate  goal is to maximize the probability of a specific (typically, ordered) sequence of actions, \( \mathbf{a}_g \),  given an initial state, i.e.

\begin{equation}
\label{eq:nolabel_5}
\max P^{\mathcal{F}, u}(\mathbf{a}=\mathbf{a}_g|c)
\end{equation}

obtained over \( \mathcal{F} \), \( s_0(c) \)  and \( u \)  where \( \mathcal{F} \)  represents some functional operation over some function that maps input tokens to output tokens (i.e. an LLM). It may include some hidden/internal state and the initial state is defined from the input context, \( c \) , and some parameterization \( \alpha, \beta, ... \), e.g.

\begin{equation}
\label{eq:nolabel_6}
s_0 = f(c, \alpha, \beta, ...).
\end{equation}

For example \( s_0 \)  may be an initial prompt template with the template variables replaced by the context, which is a common solution, while \( \mathcal{F} \) is simply a single LLM model, e.g. \( \mathcal{F}(m) = m \) where \( m \) indicates an LLM model.

Note that, in general, \( \mathcal{F} \)  is a functional that defines how some input maps to some output (albeit probabilistically). For example, \( \mathcal{F} \)  defines the probabilistic mapping of input to output by LLMs, prompt techniques (such as ReAct, CoT, etc.), inference algorithms, etc. In cases where it aids readability, we put in parentheses within the text what \( \mathcal{F} \)  specifically represents in such cases.

\begin{figure}[!htbp]
\centering
\includegraphics[width=9.03cm,height=6.37cm]{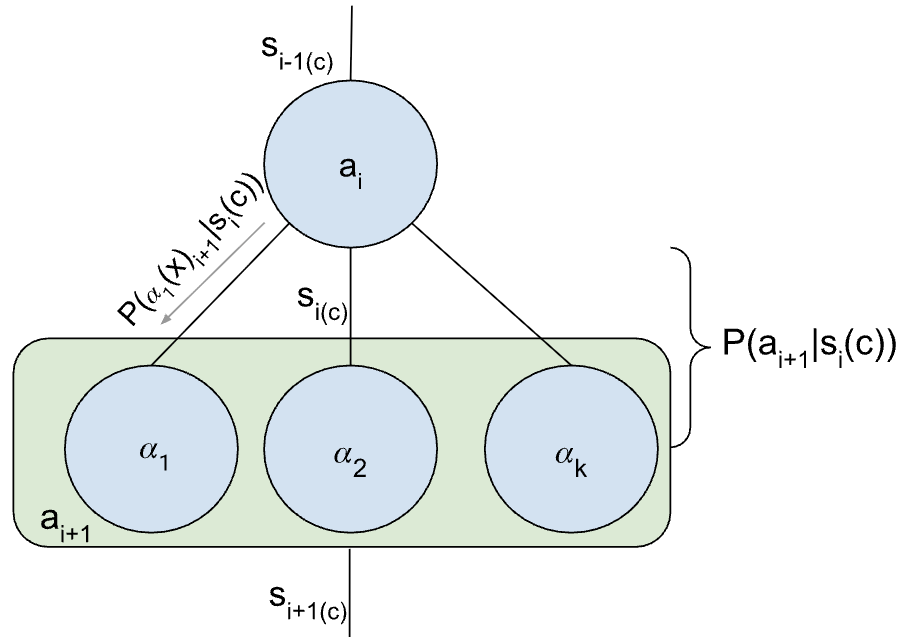}
\caption{Graphical representation of a single execution step in a ReAct-style agent loop}
\label{fig:graphical_representation_single_execution_step}
\end{figure}

\begin{figure}[!htbp]
\centering
\includegraphics[width=10.03cm,height=2.64cm]{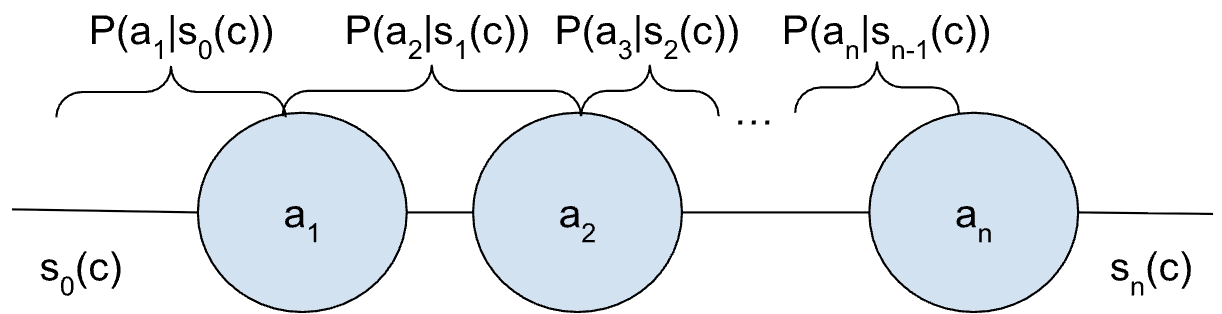}
\caption{The probability chain and state update mechanisms in a ReAct loop}
\label{fig:probability_chain_and_state_update}
\end{figure}

In the original ReAct loop, \( u \)  simply concatenates thoughts and observations, so you have two degrees of freedom, the input context (user request) and the prompt parameters (initial prompt). One step is shown as a graph in Fig. \ref{fig:graphical_representation_single_execution_step} and the whole chain is shown in Fig. \ref{fig:probability_chain_and_state_update}.

For the ReAct loop, the entire probability is a chain of the same probability kernel, \( P^{t}(a|s) \) and the state update is a concatenation of the thought and observation, e.g. \( u(a_i, s_{i-1}) = \textrm{concat}(s_{i-1}, t_i, o_i) \). The initial state is  \( s_0(c) \)  and represents the initial prompt, the tools, and user input. 

The reader should note that, here we use a Lossless Concatenation, \( u_{\textrm{concat}}(s,t,o) \) , as the update function. We recognize other update functions such as a Lossy Summarizing Update: \( u_{\textrm{summary}}(s,t,o) \)  involving summarizing \( s \), \( t \), \( o \); and a Structured Selective Update: \( u_{\textrm{selective}}(s,t,o) \)  involving a memory and a look-up approach. The choice of any of these update functions does not change the mathematical formulation.

Two ways to create more opportunities to maximize the probability of taking the desired actions are:

\begin{itemize}
\item{Use a different model for the hidden or intermediate states; not just an explicit thought but more complicated behavior, like the tree of thoughts with pruning or multiple processing steps. This is denoted by the superscript \( \mathcal{F} \)  to indicate a complex functional process.}
\item{Alternatively, one could dynamically change the state update and/or \( \mathcal{F} \)  between actions and steps, making \( u \)  and \( \mathcal{F} \)  different at each stage of the chain. This means a \textit{different} probability kernel for each transition.}
\end{itemize}

Both have a place. The first shows up in things like thinking models. The second is where approaches like different fine tuned models, using dynamic prompting, and injecting dynamic context comes into play

A third option that is a blend of these two can also be introduced. This will be shown to be a multi-agent solution - dynamically adapting the probability distribution for an action by changing the update and core probability model.

\subsection{Agent}

We now introduce a notation for an agent as

\begin{equation}
\label{eq:nolabel_7}
A^{\mathcal{F},u}(\mathbf{a}|c) = P^{\mathcal{F},u}(\mathbf{a}|c)
\end{equation}

For model functional \( \mathcal{F} \), update function \( u \), action set \( \mathbf{a} \)  and input context \( c \), used to produce state \( s(c) \).

We can partition the full action space, \( \mathbf{a} \) \textbf{,} into parts; for illustration we use three: \( \mathbf{a}_U \times \mathbf{a}_K \times \mathbf{a}_L \) . Here \( \mathbf{a}_K \) represents the set of actions \( A^{\mathcal{F},u}(\mathbf{a}=\mathbf{a}_K|\mathbf{c}) \)  which have a specific probability of being performed by an agent \( A^{\mathcal{F},u} \) .

By partitioning between two agents we have

\begin{equation}
\label{eq:nolabel_9}
P(\mathbf{a}_U \times \mathbf{a}_K \times \mathbf{a}_L|s_0(c)) = \\
\int dc_L dc_K P^{\mathcal{F}_1, u_1}(\mathbf{a}_U|s(c_K)) P(c_K|\mathbf{a}_K) P^{\mathcal{F}_2, u_2}(\mathbf{a}_K|s(c_L)) P(c_L|\mathbf{a}_L) P^{\mathcal{F}_1, u_1}(\mathbf{a}_L|s_0(c))
\end{equation}

Which can be seen as

\begin{equation}
\label{eq:nolabel_10}
\int dc_L dc_K A^{\mathcal{F}_1, u_1}(\mathbf{a}_U|s(c_K)) P(c_K|\mathbf{a}_K) A^{\mathcal{F}_2, u_2}(\mathbf{a}_K|s(c_L)) P(c_L|\mathbf{a}_L) A^{\mathcal{F}_1, u_1}(\mathbf{a}_L|s(c))
\end{equation}

\begin{figure}[!htbp]
\centering
\vspace*{5mm} 
\begin{overpic}[width=12.52cm]{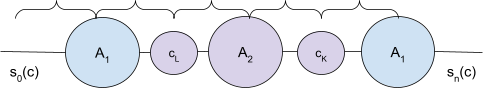}
        \put(2,23){$A^{\mathcal{F}_1,u_1}(\mathbf{a}_L|s(c))$}
        \put(21,19){$P(c_L|\mathbf{a}_L)$}
        \put(33,23){$A^{\mathcal{F}_2,u_2}(\mathbf{a}_K|s(c_L))$}
        \put(51,19){$P(c_K|\mathbf{a}_K)$}
        \put(65,23){$A^{\mathcal{F}_1,u_1}(\mathbf{a}_U|s(c_K))$}
    \end{overpic}
\caption{Graphical representation of a multi-agent interaction with a hierarchical topology}
\label{fig:graphical_representation_multiagent_interaction_hierarchical}
\end{figure}
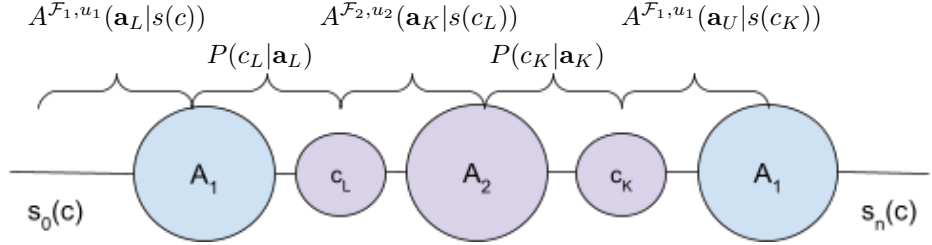

Which can be read as the probability \( \text{Agent}_1 \)  performs actions \( \mathbf{a}_L \) and uses context \( c_L \)  to task \( \text{Agent}_2 \)  to perform actions \( \mathbf{a}_K \)  which in turn leads to context \( c_K \)  which is used by \( \text{Agent}_1 \)  to perform actions \( \mathbf{a}_U \). We show this graphically in Fig. \ref{fig:graphical_representation_multiagent_interaction_hierarchical}.

Note that the partitioning here (involving \( \text{Agent}_1 \rightarrow \text{Agent}_2 \rightarrow \text{Agent}_1 \); see the figure below) captures one of the common topologies of inter-agent interactions, namely a Hierarchical Topology. For brevity, we leave out the discussions of other topologies.

Of particular note is the introduction of the following part (shown in purple in Fig. \ref{fig:graphical_representation_multiagent_interaction_hierarchical}).

\begin{equation}
\label{eq:nolabel_11}
P(c_K|\mathbf{a}_K) A^{\mathcal{F}_2, u_2}(\mathbf{a}_K|s(c_L)) P(c_L|\mathbf{a}_L).
\end{equation}

This can be read as the probability agent \( A_2 \)  returns some context \( c_K \), given it took actions \( \mathbf{a}_K \)  times the probability it took actions \( \mathbf{a}_K \)  given the state created from context \( c_L \), times the probability of context \( c_L \)  coming from the actions \( \mathbf{a}_L \). These new probabilities \( P(c_K|\mathbf{a}_K) \)  and \( P(c_L|\mathbf{a}_L) \)  are where new opportunities arise.

Considering the behavior of the new \( c_L \)  and \( c_K \)  components - in many ways these act in a similar manner to the thoughts of ReAct. But we haven't defined what \( P(c_L|\mathbf{a}_L) \)  might look like. This is where collaboration can shine, even after accounting for collaboration costs (see the Supplementary Information).

\section{Optimization Levers for Maximizing Goal Probability}

Recall the goal is to maximize the probability that specific actions are taken given the fixed parameters (model training, agent definition, tool structure, etc) by trying to change \( s \) , \( u \)  and \( \mathcal{F} \) . But now we have a new mechanism that can be used to improve our odds of the correct outcome, \( P(c_L|\mathbf{a}_L) \)  and \( P(c_K|\mathbf{a}_K) \) . 

If we can enable the two systems to search together over some space, we can more easily determine the right \( c_L \)  to provide the agent, and it, in turn, can know the right \( c_K \)  to return to the caller. This represents collaboration and negotiation!

Thought of in another way, a trace of a single execution represents a single choice of a set of hidden variables within the larger probability space - i.e. choices. Given the ability to sample infinitely many times, you would be able to map this space and determine the ideal set of choices to make. With this information you could fix \( s(c) \), \( u \)  and \( \mathcal{F} \)  such that you have the map \( c \rightarrow a \)  with the maximum possible probability, over all possible sets of \( c \)  and \( a \) .

In reality, we can't explore all possible choices. Instead we can use a new mechanism to probabilistically try to find a higher probability space. So if agent A can be probed about what context \( c_L \)  might do, without actually executing the actions, it is possible to search the subspace for a locally ‘optimal’ solution (optimal is in quotes as it is not formally optimal, but represents some improved result).

So imagine an optimal \( c_L \)  such that \textbf{(a)} you have maximized \( A^{\mathcal{F}_2, u_2}(\mathbf{a}_K|s(c_L)) \)  for your target \( \mathbf{a}_K \) and \textbf{(b)} produced the optimal \( c_K \)  to maximize \( A^{\mathcal{F}_1, u_1}(\mathbf{a}_U|s(c_K)) \) . You would be able to dynamically tune your outcomes without needing deep model training. Instead you would reach your goals (as maximally as possible) by dynamically searching!

Of course, this is not a reality. A maximization like this is not feasible, and even if it were, a maximized probability does not mean a high probability. It just means given the domain of input variables and the constraints to achieve this is the maximal probability possible. But what this approach does do is the following:

\begin{itemize}
	\item Partitions the action space and allows for increased probability within a subspace 

	\item Allows for a search algorithm to be developed that maximizes the likelihood of the desired outcomes, without needing to change the inputs

\end{itemize}
Partitions of the actions are critically important. As a probability is always bound between 0 and 1, as we chain events together we are always, at best, maintaining the probability of the desired outcome at the level it is after the first i actions. If the number of needed actions grows, the probability of end to end success shrinks. Similarly as the number of possible actions increases, the probability of choosing the right action at each step goes down.

By reducing the number of possible actions and the length of action sequences, you give yourself the best chance at success. Thus, by encapsulation of a partition of actions into an agent that has tuned \( \mathcal{F} \), \( u \)  and \( s(c) \), you have a higher probability of success on the subactions \( \mathbf{a}_K \)  performed by the agent, than is likely by trying to make these decisions within the same constrained \( \mathcal{F} \), \( u \)  and \( s \) that the rest of your actions are tied to.

Taken to the extreme, you would be most likely to succeed if you could dynamically explore every link between action \textbf{a} and action \textbf{b}. But the reality is that most real actions (aka tools or LLM processing) do not allow probing. Additionally, probing might be expensive itself, practically. Thus the best approach is to isolate some sub actions to an agent that can be probed to dynamically maximize performance characteristics per request.

\section{Relationship to Standards}

Where does Model Context Protocol (MCP) fit in this picture? MCP, practically, provides a set of tools \( \alpha \), a standard syntax to shape how arguments, \( x \), are produced by a model as well as support for prompt templates to use in shaping the state consumed by the model  \cite{Aditi_Singh_et_al_2025}. This allows models to be trained to improve the probability of \( P^{\mathcal{F},u}(a_i|s_{i-1}) \), by improving odds with changes to \( u \)  and \( s_{i-1} \).

Agent-to-Agent Protocol (A2A)  \cite{Partha_Pratim_Ray2025}, on the other hand, provides a pathway to building an algorithm to optimize for \( c_L \)  as well as a medium for \( c_L \)  to be passed, \( \mathbf{a}_K \)  to be executed and \( c_K \)  to be returned. By allowing agents to be developed with improved odds by changing \( \mathcal{F} \) , \( u \)  and \( s \) , A2A allows for easier partitioning of the action space in a way likely to lead to higher probability outcomes.

\section{Different Strategies and Paradigms}

Now let's walk through some examples and how they relate to this formalism.

\begin{figure}[!htbp]
\centering
\includegraphics[width=12.18cm,height=8.47cm]{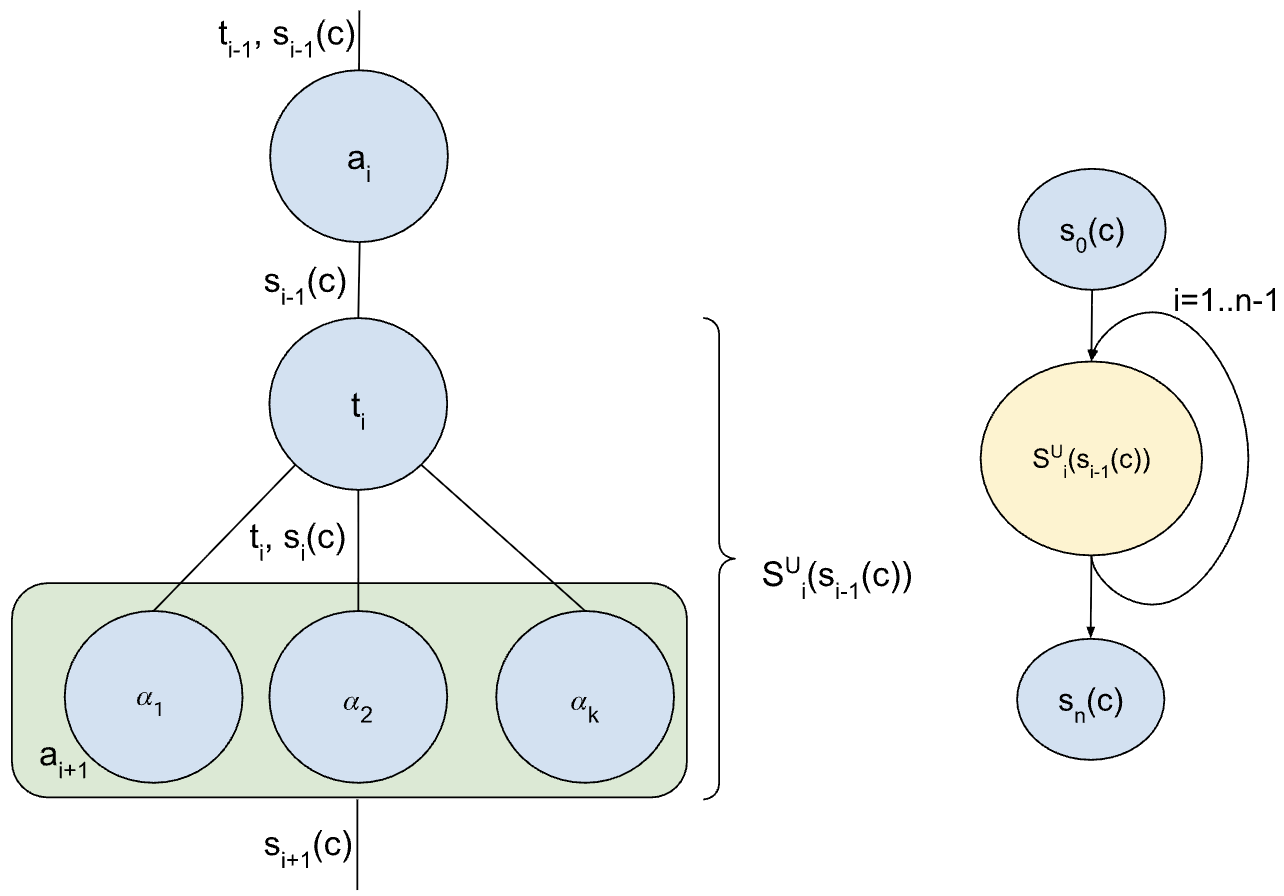}
\caption{Visualization of the ReAct strategy and its limitations}
\label{fig:visualization_react_strategy_and_its}
\end{figure}

We discussed ReAct, and its relationship. For ReAct, the only knobs you have, generally speaking, are the state update, the original state creation (from a user prompt) and the set of possible actions, \( \alpha \)  that can be taken. See Fig. \ref{fig:visualization_react_strategy_and_its}.

The drawbacks to this approach are that it can't, easily, apply more constraints. You can include them in your prompt, e.g.the state that drives the probability of an action, but you can’t enforce it. A more advanced ReAct loop might include logic to remove possible actions, \( \{\alpha\} \)  under certain conditions, which starts to move away from the core ReAct loop and becomes a smart graph editing algorithm. Additionally, ReAct, which is fundamentally a random walk, suffers from other issues due to lack of controls, like constraint enforcement and lack of convergence. The lack of convergence is often manifested as hallucinations which lead the random walk away from the desired state.

An alternative approach would be to build a richer system that maps a state to an LLM prompt, enhancing the probability of a certain outcome. This is what might happen with a richer contextual management layer. Here, you build some model to improve the odds of \( P^{\mathcal{F}}(\mathbf{a}|\mathbf{c}) \)  by tweaking \( s(c) \) . The user input, \( c \) , remains unchanged, but it is used in a more complicated way to create the state (essentially LLM context) \( s(c) \) .

Another approach is to develop a new processing functional, \( \mathcal{F} \), that improves the odds of the correct outcome. For example, a deep thinking model uses a more complicated set of LLM inferences to drive improved results for logical inference. Similarly there are numerous prompting techniques that all propose a different effective \( \mathcal{F} \). 

Yet another approach might be to partition and organize actions so that a certain ordering or sequence is adhered to. This way you can increase the probability of specific actions at a certain point in the sequence by manipulating the allowed actions. This approach greatly improves the probability of specific actions by radically reducing the options based on a specific control flow. This is another form of manipulating \( \mathcal{F} \), \( u \)  and \( s(c) \)  at each step in the sequence to increase the desired outcome at each step. This is how a graph based system would operate. At each node in the graph a different \( s(c) \)  is utilized, with a different \( u \)  and possibly different \( \mathcal{F} \).  This allows the developer to constrain the action space and implicitly constrain the context space (by virtue of only certain paths can arrive at a certain node).

This approach of partitioning can be used to also help understand parallelism. In parallelism, there is a set of actions that are independent of a different set of actions. Mathematically this means the state the probability is conditioned on is decoupled, i.e.

\begin{equation}
P^{\mathcal{F}_1}(\mathbf{a}_X|s_{i-1}) \times P^{\mathcal{F}_2}(\mathbf{a}_Y|s_{i-1})
\end{equation}

\begin{figure}[!htbp]
\centering
\includegraphics[width=10.41cm]{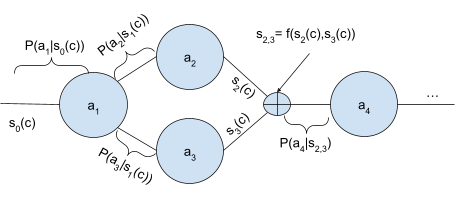}
\caption{Formalization of parallelism and recombination}
\label{fig:formalization_parallelism_and_recombination}
\end{figure}

Where actions \( \mathbf{a_X} \) and \( \mathbf{a_Y} \)  are dependent on \( s_{i-1} \) , but not on the result of either of the other probabilities. This is no longer a Markov chain, but a graph of Markov chains. Recombining the result of the parallelism is then another problem space itself. For example, taking the result of the two branches to generate some new actions would look like:

\begin{equation}
\label{eq:nolabel_13}
P^{\mathcal{F}}(\mathbf{a}_N | s(s_{i-1,X}, s_{i-1,Y})) P^{\mathcal{F}_1}(\mathbf{a}_X|s_{i-1}) \times P^{\mathcal{F}_2}(\mathbf{a}_Y|s_{i-1})
\end{equation}

with \( s(s_{i-1,X}, s_{i-1,Y}) \)  representing the new combination logic for taking the result of the actions \( \mathbf{a_X} \)  on \( s_{i-1} \)  and \( \mathbf{a_Y} \)  on \( s_{i-1} \)  and combining them to a new state. This type of splitting can be shown graphically as illustrated in Fig. \ref{fig:formalization_parallelism_and_recombination}.

This general approach, and the resulting problem of recombination, holds true whether the branch represents a single action, or a full agentic set of actions. With agentic actions, we also have the ability to introduce \( P(c_X | \mathbf{a}_X) \)  and \( P(c_Y | \mathbf{a}_Y) \)  for the agentic nodes (collaboration and negotiation).

\section{Methodology Comparison and Degrees of Freedom}

The rapid growth of diverse agent architectures necessitates a clear and unified method for comparison. We formalize that comparison by explicitly mapping various agent strategies and prompting methodologies to the probabilistic framework introduced earlier. It serves to identify and differentiate the distinct "Degrees of Freedom" – the optimizable parameters and functional levers – available to the agent designer. The core objective is to move beyond empirical intuition by providing a clear understanding of the architectural trade-offs. The following conceptual terms are fundamental to this analysis:

\begin{itemize}
\item{\textbf{Prompt Engineering:} The static manipulation of the initial state, \( s_0(c) \).}
\item{\textbf{Context Engineering:} The dynamic, strategic manipulation of the state \( s_i(c) \)  at each step.}
\item{\textbf{Inference Algorithms:} The definition of fixed patterns for the inference functional \( \mathcal{F} \)  and state update function \( u(a_i, a_{i-1},..., a_1, s_{i-1}, s_{i-1},..., s_0) \).}
\end{itemize}

We summarize in Table 1 how these principles manifest as optimizable parameters across different agent methodologies, from the monolithic ReAct loop to complex Multi-Agent Systems with collaboration.

\begin{table}[!htbp]
\renewcommand{\arraystretch}{1.3}
\begin{adjustbox}{max width=\textwidth}
\begin{tabular}{p{3.33cm}p{8.02cm}p{6.96cm}p{3.33cm}p{8.02cm}p{6.96cm}}
\hline
\multicolumn{1}{|p{3.33cm}}{Methodology} & 
\multicolumn{1}{|p{8.02cm}}{Degrees of Freedom} & 
\multicolumn{1}{|p{6.96cm}|}{Optimizing Parameters} \\ 
\hline
\multicolumn{1}{|p{3.33cm}}{ReAct} & 
\multicolumn{1}{|p{8.02cm}}{Prompt Engineering, State Update} & 
\multicolumn{1}{|p{6.96cm}|}{\( s_0(c) \) , \( u(a_i, s_{i-1}) \) } \\ 
\hline
\multicolumn{1}{|p{3.33cm}}{Action-based Composable LLM Inference} & 
\multicolumn{1}{|p{8.02cm}}{Inference Graph} & 
\multicolumn{1}{|p{6.96cm}|}{\( \mathcal{F} \) } \\ 
\hline
\multicolumn{1}{|p{3.33cm}}{Deep Thinking} & 
\multicolumn{1}{|p{8.02cm}}{Inference Graph, Model parameters} & 
\multicolumn{1}{|p{6.96cm}|}{\( \mathcal{F} \) } \\ 
\hline
\multicolumn{1}{|p{3.33cm}}{Fine Tuning} & 
\multicolumn{1}{|p{8.02cm}}{Model parameters} & 
\multicolumn{1}{|p{6.96cm}|}{\( \mathcal{F} \) } \\ 
\hline
\multicolumn{1}{|p{3.33cm}}{Control Flow} & 
\multicolumn{1}{|p{8.02cm}}{Partition actions, change prompts per step, state update. Support non-Markov chain use cases} & 
\multicolumn{1}{|p{6.96cm}|}{\( \mathcal{F} \) \textit{, \( u(a_i, s_{i-1}) \) , \( s_0(c) \) , \( a_i \rightarrow \{\alpha_i, x_i, o_i\} \)   }} \\ 
\hline
\multicolumn{1}{|p{3.33cm}}{Multi-Agent (no collaboration)} & 
\multicolumn{1}{|p{8.02cm}}{Partition actions and agent logic. Support non-Markov chain use cases} & 
\multicolumn{1}{|p{6.96cm}|}{\( \mathcal{F} \) \textit{, \( u(a_i, s_{i-1}) \) , \( s_0(c) \) , \( a_i \rightarrow \{\alpha_i, x_i, o_i\} \)   }} \\ 
\hline
\multicolumn{1}{|p{3.33cm}}{Multi-Agent (collaboration)} & 
\multicolumn{1}{|p{8.02cm}}{Partition actions and agent logic, enable collaboration and negotiation. Support non-Markov chain use cases} & 
\multicolumn{1}{|p{6.96cm}|}{\( \mathcal{F} \) \textit{, \( u(a_i, s_{i-1}) \) , \( s_0(c) \) , \( a_i \rightarrow \{\alpha_i, x_i, o_i\} \) , \( P(c_L|\mathbf{a}_L) \) , \( P(c_K|\mathbf{a}_K) \) }} \\ 
\hline
\end{tabular}
\end{adjustbox}
\caption{Degrees of Freedom and Optimizing Parameters for Agent Architectures}
\label{tab:degrees_freedom_and_optimizing_parameters}\end{table}

\begin{figure}[!htbp]
\centering
\includegraphics[width=10.41cm]{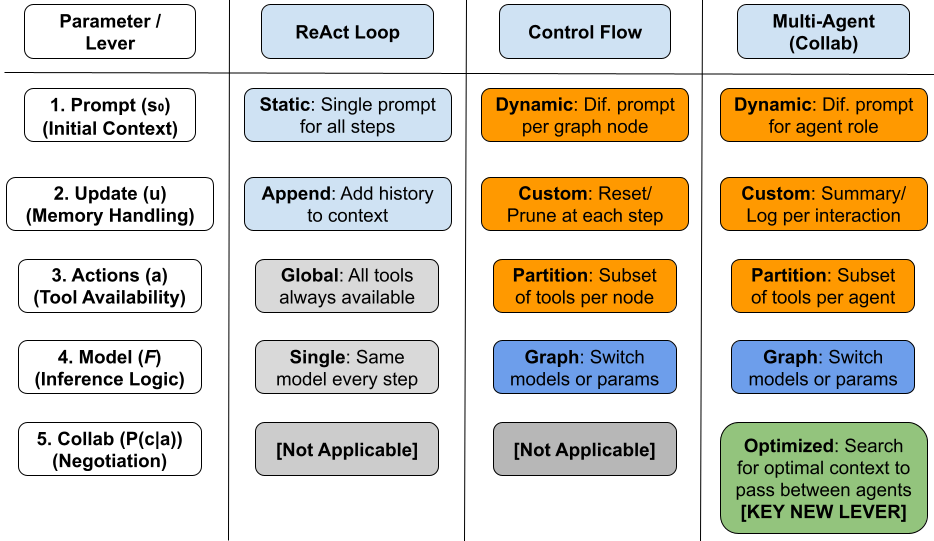}
\caption{The Strategy Capability Board. We contrast the "locked" nature of monolithic architectures against the expanded degrees of freedom in Control Flow and Multi-Agent systems. Note that the "Collaboration" lever (optimizing $P(c|a)$) is unique to the multi-agent paradigm.}
\label{fig:strategy_board}
\end{figure}

To provide a more intuitive understanding of these architectural distinctions, we visualize the "Degrees of Freedom" as a set of operational levers in Fig. \ref{fig:strategy_board}. This "Strategy Capability Board" illustrates the progression of optimizable parameters from the monolithic ReAct loop to collaborative Multi-Agent systems.
While the ReAct framework is characterized by static prompts ($s_0$) and global action spaces ($a$), Control Flow architectures "unlock" the ability to partition these spaces per step, effectively allowing for dynamic prompting and tool restriction at each node in the graph. Uniquely, the Multi-Agent paradigm activates the "Collaboration" lever – mathematically represented by the probability $P(c_L|a_L)$ – transforming inter-agent communication from a fixed protocol into a dynamically optimizable search space. Overall, Figure 7 further illustrates how the transition to multi-agent architectures is not merely a structural change, but an expansion of the probabilistic surface area available for optimization.

\section{Conclusion}

The rise of sophisticated AI agents – from monolithic ReAct loops to complex multi-agent swarms – has created a critical and challenging divergence in design philosophy. This paper successfully addresses the fundamental challenge: the lack of a unified, rigorous language to formally describe, analyze, and compare these disparate strategies.

We have established that the behavior of any agentic system can be fundamentally understood as a probabilistic process. By framing the agent's operation as a probability chain aiming to maximize the probability of a goal action sequence, \( P(\mathbf{a}=\mathbf{a}_g|c) \) , we have provided a common mathematical context for all agent architectures.

Our key contributions – the \textbf{Unified Probabilistic Formulation}, the intuitive \textbf{"Degrees of Freedom"} concept, and the \textbf{Formalization of Collaboration} – serve to transition agent design from an empirical art to a systematic engineering discipline. Specifically, we demonstrated how the multi-agent paradigm introduces powerful new degrees of freedom, such as the inter-agent context probabilities \( P(c_L|\mathbf{a}_L) \)  and \( P(c_K|\mathbf{a}_K) \) , which act as optimizable levers for dynamic, runtime tuning of the outcome probability. Furthermore, by introducing a regularized objective function that accounts for \textbf{Collaboration Costs}, we offer a more robust framework that guides designers toward solutions that are both theoretically effective and practically efficient.

In essence, this work provides the intellectual bridge connecting high-level agent concepts with precise, quantifiable principles. It offers designers a clear roadmap for dissecting and optimizing architectural trade-offs, enabling the systematic maximization of success probability within any complex agentic system. Future research can leverage this framework to develop sophisticated, probabilistically-grounded search algorithms that dynamically exploit these new degrees of freedom, accelerating the journey toward reliable and high-performing AI Systems.

\section{Supplementary Information}

\subsection{Accounting for Collaboration Costs}

While multi-agent collaboration introduces powerful new mechanisms to optimize outcomes, as represented by the probabilities \( P(c_L|\mathbf{a}_L) \)  and \( P(c_K|\mathbf{a}_K) \) , this added capability is not without potential trade-offs. Real-world implementations of inter-agent communication and negotiation inherently incur costs, which can manifest as increased latency, higher computational or token usage, and greater system complexity. Ignoring these factors could lead to a theoretical model that overestimates the practical benefits of collaboration.

To create a more robust and realistic objective function, we can introduce a regularization term to account for these "Collaboration Costs." This term acts as a penalty, tempering the drive to maximize the probability of the goal actions by factoring in the resources consumed during collaboration. This ensures that the selected strategy is not just effective in terms of outcome probability but is also efficient.

We can modify our original objective function to include this new component. Instead of purely seeking to maximize the probability of the goal action sequence \( \mathbf{a}_g \) , we now aim to maximize a cost-adjusted probability:

\begin{equation}
\label{eq:nolabel_15}
\text{Maximize} \left( P(...) - \lambda \cdot \text{CollabCost}(\mathbf{a}_g, c, \mathcal{F}, u) \right)
\end{equation}

which expands to \(  \text{Maximize} \left( P_{\mathcal{F},u}(\mathbf{a} = \mathbf{a}_g | c) - \lambda \cdot \text{CollabCost}(\mathbf{a}_g, c, \mathcal{F}, u) \right)  \) 

Here, \( \text{CollabCost}(\cdot) \)  represents a function that quantifies the accumulated costs (latency, tokens, complexity) associated with the collaborative steps within the action sequence \( \mathbf{a}_g \) . The term \( \lambda \)  is a small, positive hyperparameter that weighs the importance of this cost. A higher \( \lambda \)  would more strongly penalize costly collaborations, steering the system towards simpler, faster, or less resource-intensive agent interactions.

This revised objective, \( \text{Maximize}( P(...) - \lambda \cdot \text{CollabCost}(\cdot) ) \) , provides a more complete framework. It allows the system to find a balance, favoring agent architectures that yield a high probability of success while remaining efficient. This regularization helps address practical concerns and guides the design toward solutions that are not just theoretically optimal but also viable in real-world applications where resources are finite.
\bibliographystyle{plain}
\bibliography{bibliography-bibtex.bib}

\end{document}